\documentclass{article}
\usepackage{spconf,amsmath,graphicx}

\usepackage{enumitem}
\setlist{nosep, leftmargin=14pt}

\usepackage{mwe} % to get dummy images
\usepackage{color}
\usepackage[table]{xcolor}
\usepackage{amsmath}
\usepackage{amssymb}
\usepackage{multirow}
\usepackage{color}
\usepackage{bm}
\usepackage{bbm}
\usepackage{arydshln} 
\usepackage{mathtools} % 数式表示用パッケージ
\usepackage{hyperref}
\hypersetup{%
 colorlinks=true}
\title{Hypernetwork-Based Adaptive Aggregation for Multimodal Multiple-Instance Learning in Predicting Coronary Calcium Debulking}
\name{Kaito Shiku$^{1 *}$, Ichika Seo$^{1 *}$, Tetsuya Matoba$^{2}$, Rissei Hino$^{2}$, Yasuhiro Nakano$^{2}$, Ryoma Bise$^{1}$\thanks{$^{*}$ Contributed equally.}}
\address{$^{1}$ Department of Advanced Information Technology, Kyushu University \\ $^{2}$ Department of Cardiovascular Medicine, Kyushu University}
\begin{document}
% \ninept
%
\maketitle
\begin{abstract}
In this paper, we present the first attempt to estimate the necessity of debulking coronary artery calcifications from computed tomography (CT) images.
We formulate this task as a Multiple-instance Learning (MIL) problem.
The difficulty of this task lies in that physicians adjust their focus and decision criteria for device usage according to tabular data representing each patient’s condition.
To address this issue, we propose a hypernetwork-based adaptive aggregation transformer (HyperAdAgFormer), which adaptively modifies the feature aggregation strategy for each patient based on tabular data through a hypernetwork.
The experiments using the clinical dataset demonstrated the effectiveness of HyperAdAgFormer.
The code is publicly available at \url{https://github.com/Shiku-Kaito/HyperAdAgFormer}.
\end{abstract}
%
% \begin{keywords}
% Multimodal Multiple-instance Learning, Hypernetworks, Adaptive Aggregation
% \end{keywords}
%
\section{Introduction}
\label{sec:intro}
Coronary artery calcification (CAC) is a major cause of vascular hardening, stenosis, and impaired coronary blood flow~\cite{onnis2024coronary}. 
In percutaneous coronary intervention (PCI)~\cite{chan2011appropriateness}, a stenotic segment is dilated by advancing a catheter and inflating a balloon. 
However, when calcification is severe, the catheter cannot cross the lesion and the balloon fails to expand the lumen. 
In such cases, a debulking device is used to modify or shave the calcified plaque~\cite{barbato2017state}. 
At present, the necessity of these devices is determined intraoperatively based on the vascular and patient conditions, which prolongs the procedure due to device preparation and imposes additional burden on the patient~\cite{bacmeister2023planned}. 
On the other hand, computed tomography (CT) images can be acquired preoperatively without imposing any additional burden on the patient. Therefore, if the need for a debulking device can be estimated preoperatively from CT images, patient burden could be reduced, procedural preparation streamlined, and procedural success rates improved. Accordingly, developing an automated method to predict the necessity of debulking device use from preoperative CT images is highly desirable.

% At present, the necessity of these devices is determined intraoperatively based on the vascular and patient conditions, which prolongs the procedure due to device preparation and imposes additional burden on the patient~\cite{bacmeister2023planned}. 
% If the need for a debulking device could be estimated preoperatively from computed tomography (CT) images, patient burden could be reduced, preparation streamlined, and procedural success rates improved.
% Therefore, developing an automated method that can predict the necessity of using debulking device from preoperative CT images is highly desirable.

The goal of this paper is to develop a model for estimating whether a debulking device is required based on CT images.
To the best of our knowledge, this is the first study to tackle this problem using preoperative imaging data.
We leverage diagnostic data that are routinely collected and extensively stored during daily clinical practice to train the model.
The available surgical record data include only CT images and information on whether a debulking device was used; there are no records indicating which specific calcified regions were treated with the device.

We formulate this task as a Multiple-instance Learning (MIL) problem~\cite{ramon2000multi, ilse2018attention, shao2021transmil, wang2018revisiting, Shiku_2025_WACV, kaito2025learning}, where each cross-sectional image of the vessel in the CT image is treated as an instance, and the entire vessel of each patient, represented by a set of slice images, is regarded as a bag for classifying the necessity of using the device.
Here, cross-sectional images of the coronary vessel can be readily obtained using commercially available vessel extraction software commonly used in clinical practice.
In this setting, annotations indicating which calcifications require debulking are unavailable; only patient-level surgical records indicating whether a device was used serve as supervision for the corresponding set of images.

\begin{figure}
      \centering
        \includegraphics[width=0.93\linewidth]{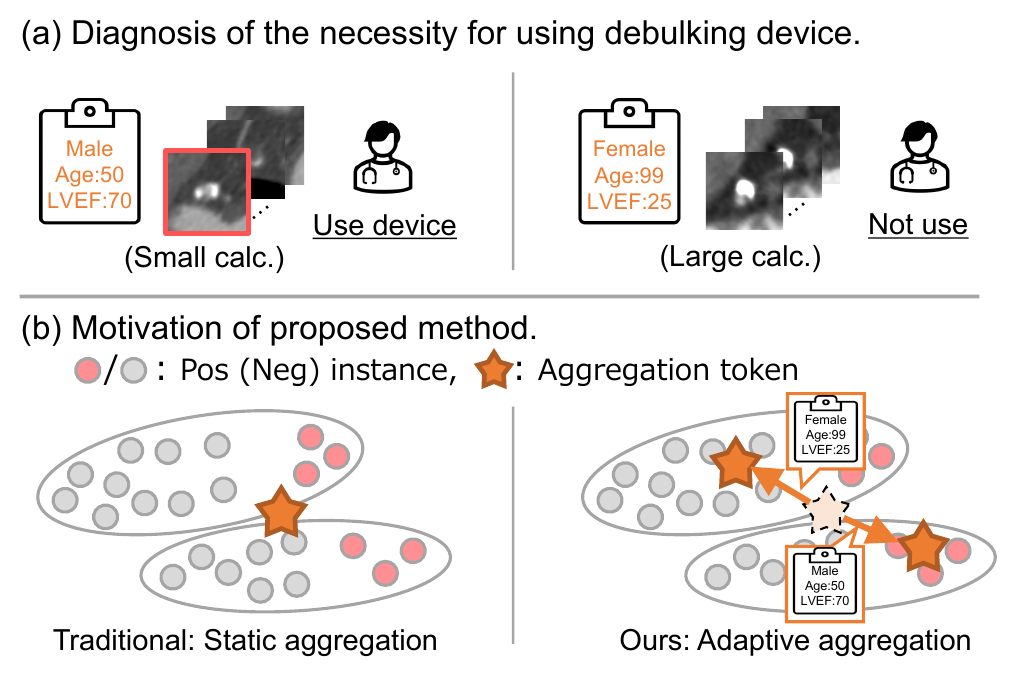}
        \vspace{-4mm}
        \caption{(a) Clinical diagnosis of the necessity for using debulking device. (b) Motivation of the proposed tabular data–based adaptive aggregation.
        \label{fig:intro}}
        \vspace{-4mm}
\end{figure}

In clinical practice, surgeons determine the necessity of using a debulking device based on both CT findings and patient-specific clinical information such as age, cardiac function, and surgical risk. As shown in Fig.~\ref{fig:intro}(a), surgeons tend to treat even small calcifications in young, low-risk patients, whereas in elderly, high-risk patients they focus only on larger or more severe ones. These differences indicate that regions of interest vary with patient conditions, motivating an MIL approach that adaptively adjusts its focus regions according to tabular data.

These observations suggest that the decision process inherently differs across patients, indicating the need for a model that can adapt its attention according to patient-specific information. 
However, existing MIL methods are typically designed for unimodal image data and employ a single, static attention mechanism optimized over all training samples.
Several attention-based aggregation methods have been proposed in unimodal MIL studies~\cite{ilse2018attention, shao2021transmil, Shiku_2025_WACV}, where the aggregation focuses on the most informative instances.
As shown in Fig.~\ref{fig:intro}(b), such static attention strategies cannot achieve patient-specific adaptive aggregation, which is essential for our task.
A few MIL studies have incorporated tabular data~\cite{li2021multi, xu2023multimodal, hemker2024healnet}, typically aiming to learn dense correlations between modalities. 
In our task, however, the tabular features—age, sex, LVEF, and HF—reflect patient-level conditions rather than instance-level patterns, so simple feature fusion may fail to capture their complementary influence on image-based prediction.

% To address these challenges, we propose a hypernetwork-based adaptive aggregation transformer (HyperAdAgFormer) for multimodal MIL, which adaptively modifies the feature aggregation strategy for each patient based on tabular data through a hypernetwork.
% A hypernetwork is a model that generates a subset of parameters for a primary network responsible for a specific task.
% In HyperAdAgFormer, tabular data are used as input to estimate the Tabular-conditioned Transformation Parameters (TCTP) for the aggregation token of the transformer model, which performs instance feature aggregation, and to generate the bag-level classifier.
% As shown in Fig.~\ref{fig:intro}(b), by applying this transformation to the aggregation token for each patient, the model achieves adaptive aggregation tailored to individual patients based on their clinical information.

To overcome these limitations, we propose a hypernetwork-based adaptive aggregation transformer (HyperAdAgFormer) for multimodal MIL. 
The key idea is to achieve patient-specific adaptive aggregation by conditioning the transformer's attention on tabular clinical data. 
The hypernetwork takes the tabular data as input and generates Tabular-conditioned Transformation Parameters (TCTP) to modify the aggregation token and produce the bag-level classifier. 
As shown in Fig.~\ref{fig:intro}(b), this enables the model to perform patient-specific feature aggregation tailored to individual clinical profiles.
% 6. 実験
In experiments using clinical CT and tabular datasets, the proposed approach demonstrates its effectiveness.

% \vspace{90mm}

\section{Related work}
% \noindent
% {\bf CT images analysis.}
% % A wide range of CT image analysis tasks using machine learning models have been actively studied, such as organ and tumor segmentation~\cite{liu2024learning, lai2024pixel}, vessel tracing~\cite{zhou2023henet}, and calcification detection~\cite{jiang2024iarcac, jiang2025coronary}.
% % On the other hand, although estimating the necessity of using a debulking device for coronary calcification based on CT images is a medically important task, no studies have yet addressed this problem.
% A wide range of CT image analysis tasks have been actively studied, including organ, tumor, and calcification segmentation~\cite{liu2024learning, lai2024pixel, jiang2024iarcac, jiang2025coronary}, vessel tracing~\cite{zhou2023henet}, and prognosis prediction~\cite{dong2023improved}.
% However, despite its medical importance, no studies have addressed the task of estimating the necessity of using a debulking device for coronary calcification based on CT images.

\noindent
{\bf Multiple-instance Learning (MIL).}
% Multiple Instance Learning (MIL)~\cite{ramon2000multi, ilse2018attention, schwab2022automatic, shao2021transmil, wang2018revisiting, Shiku_2025_WACV} aims to estimate the label of a bag composed of a set of instances (e.g., images).
% MIL methods are broadly divided into output aggregation, which obtains a bag-level prediction score by aggregating the prediction scores estimated for each instance by the classification model, and feature aggregation, which aggregates instance-level features extracted by the feature extraction model and performs classification on the resulting bag-level representation.
% For each scheme, aggregation strategies such as max~\cite{wang2018revisiting}, mean~\cite{Ilse2020DeepMI}, and attention-based weighted summation~\cite{ilse2018attention} have been proposed.
MIL methods are broadly divided into two categories: output aggregation~\cite{wang2018revisiting, kaito2025learning}, which obtains a bag-level prediction score by aggregating the prediction scores estimated for each instance, and feature aggregation~\cite{ilse2018attention, shao2021transmil, wang2018revisiting, Shiku_2025_WACV, Ilse2020DeepMI}, which aggregates instance features and performs classification on the resulting bag-level representation.
For each scheme, various aggregation strategies have been proposed, including max pooling, mean pooling~\cite{wang2018revisiting, Ilse2020DeepMI}, and attention-based weighted summation~\cite{ilse2018attention}.
In state-of-the-art studies, transformer-based self-attention aggregation methods have been adopted~\cite{shao2021transmil, Shiku_2025_WACV}.
However, these studies optimize a static aggregation strategy across the entire dataset, and none have explored an approach like the proposed HyperAdAgFormer, which dynamically adapts the aggregation strategy for each patient to achieve optimal aggregation.

\noindent
{\bf Multimodal fusion methods.}
Multimodal learning is a paradigm that estimates labels using data from multiple heterogeneous modalities~\cite{li2021multi, xu2023multimodal, hemker2024healnet, chen2020pathomic}.
% In recent studies, intermediate fusion has become the mainstream approach for leveraging information across modalities, where feature vectors extracted from each modality are integrated for classification.
% Various fusion strategies have been proposed, including concatenation, guided attention fusion using weighted Kronecker products~\cite{chen2020pathomic}, and cross-attention-based fusion methods~\cite{xu2023multimodal, hemker2024healnet}.
To date, several feature fusion methods have been proposed to leverage information from both modalities for classification, including simple concatenation, guided attention using weighted Kronecker products~\cite{chen2020pathomic}, and cross-attention-based approaches~\cite{xu2023multimodal, hemker2024healnet}.
Furthermore, several studies have explored the use of tabular data in MIL, primarily aiming to enable dense information sharing between instance and tabular features through concatenation-based joint aggregation~\cite{li2021multi}, transformer-based self-attention aggregation~\cite{xu2023multimodal}, and cross-attention-based correlation modeling~\cite{hemker2024healnet}.
While these approaches assume comparable levels of information between image and tabular data and aim to achieve dense information sharing, our task involves sparse tabular data that are not directly related to the image data, which limits the effectiveness of such dense fusion methods.

\noindent
{\bf Hypernetworks.}
% A hypernetwork, originally proposed by Ha et al.~\cite{ha2017hypernetworks}, is a neural network that dynamically generates the weights and biases of a primary network conditioned on the input.
% Hypernetworks have been applied to a wide range of tasks, including multi-task learning~\cite{tay2021hypergrid}, continual learning~\cite{ohs2019hypercl}, segmentation, and domain adaptation~\cite{serebro2025hyda}.
% A hypernetwork~\cite{ha2017hypernetworks} is a neural network that dynamically generates a subset of the parameters of a primary network based on the input.
A hypernetwork~\cite{ha2017hypernetworks, tay2021hypergrid, ohs2019hypercl, serebro2025hyda} is a neural network that dynamically generates a subset of the parameters of a primary network based on the input.
% It has been applied to various tasks, including multi-task learning~\cite{tay2021hypergrid}, continual learning~\cite{ohs2019hypercl}, segmentation, and domain adaptation~\cite{serebro2025hyda}.
HyperFusion~\cite{duenias2025hyperfusion} used a hypernetwork to generate parameters of an image classification model from tabular data. 
However, this approach estimates the parameters of an image-level feature extractor and cannot be directly applied to MIL, which requires extracting features from a set of images.
To the best of our knowledge, no prior work has applied hypernetworks to enhance feature aggregation in MIL.

% Duenias et al.~\cite{duenias2025hyperfusion} used a hypernetwork to generate the parameters of a medical image classification model based on tabular data. However, their approach focuses on estimating the parameters of a feature extractor for image-level classification and cannot be directly applied to MIL, which requires aggregating features from multiple images.
% To the best of our knowledge, no prior work has applied hypernetworks to enhance feature aggregation in MIL.
% Duenias et al.~\cite{duenias2025hyperfusion} used a hypernetwork to generate the parameters of a medical image classification model based on tabular data.
% However, this approach estimates the parameters of an image-level classification model and cannot be directly applied to MIL, which requires aggregating features from multiple images.
% To the best of our knowledge, no prior work has applied hypernetworks to enhance feature aggregation in MIL.

\begin{figure}
      \centering
        \includegraphics[width=0.86\linewidth]{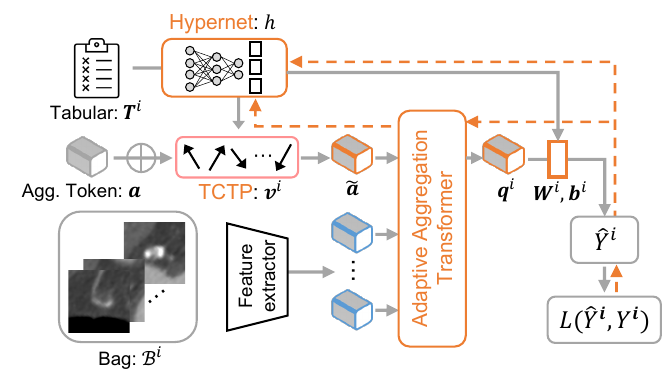}
        \vspace{-4mm}
        \caption{{\bf Overview of Hypernetwork-based Adaptive Aggregation Transformer (HyperAdAgFormer).} The gray arrows indicate the inference flow, while the orange dashed arrows represent the gradient flow to the hypernetwork.  
        \vspace{-4mm}
        \label{fig:HyperAdAgFormer}}
\end{figure}

\section{Multimodal MIL for debulking device necessity prediction}
% for estimating the necessity of using debulking devices.}
\label{sec:print}
\subsection{Problem setup}
Given the $i$-th bag (i.e., patient) consisting of instances (i.e., coronary cross-sectional slices) $\mathcal{B}^i = \{\bm{x}_{j}^i \}_{j=1}^{|\mathcal{B}^i|}$ and corresponding tabular data $\bm{T}^i$, our objective is to estimate the necessity of using a debulking device $Y^i \in \{0, 1\}$, where 1 indicates ``use'', 0: ``not use''.
Here, $|\mathcal{B}^i|$ represents the total number of instances, and $\bm{T}^i$ is a vector with $k$ values representing the patient’s condition (e.g., age, sex, HF, LVEF).
% Note that no detailed information is provided at the instance level, such as whether an instance contains calcification or whether the use of a debulking device is required.
Note that no detailed information is provided at the instance level, such as whether the use of a debulking device is required.

\subsection{Preliminary: transformer base aggregation}
In general transformer-based MIL methods, the instance features within a bag, $\{\bm{e}_{j}^i \}_{j=1}^{|\mathcal{B}^i|}$, are input to the transformer along with an aggregation token $\bm{a}$, and the bag-level feature $\bm{q}$ is obtained through self-attention, as follows:
\begin{equation}
\bm{q}^i = \text{Transformer}(\bm{a}, \bm{e}_{1}^i, \ldots, \bm{e}_{|\mathcal{B}^i|}^i).
\end{equation}
In these methods, the aggregation token is treated as a learnable parameter within the  model and is optimized so that a single shared token can estimate higher attention scores for important instances across all bags in the training data.

However, in our task, the instances important for prediction vary with each patient’s condition. Thus, rather than optimizing a shared aggregation token, it is crucial to generate an optimal one for each patient based on individual information.

\subsection{Hypernetwork-based Adaptive Aggregation Transformer}
Fig.~\ref{fig:HyperAdAgFormer} illustrates the overview of the proposed Hypernetwork-based Adaptive Aggregation Transformer (HyperAdAgFormer), which is designed to perform patient-specific adaptive feature aggregation based on tabular data representing patient information during both training and inference.

The HyperAdAgFormer consists of two main steps: estimation of the Tabular-conditioned Transformation Parameters (TCTP) by the hypernetwork, and adaptive aggregation of instance features using the aggregation token transformed by the TCTP.
The details of each step are described below.

\noindent
{\bf Tabular-conditioned Transformation Parameters Estimation by hypernetwork.}
% In this step, we aim to obtain the Tabular-Conditioned Transformation Parameters (TCTP), which adaptively transform the aggregation token for each patient, as well as the corresponding bag-level classifier parameters, by inputting the tabular data representing each patient’s information into the hypernetwork.
% Given the tabular data $\bm{T}^i$, the hypernetwork $h$ estimates the TCTP $\bm{v}^i$ and the classifier’s weights and biases $(\bm{W}^i, \bm{b}^i)$ as follows:
% \begin{equation}
%     \bm{v}^i, \bm{W}^i, \bm{b}^i  = h(\bm{T}^i).
% \end{equation}
% Here, $\bm{v}^i$ is a $d$-dimensional vector with the same dimension as the aggregation token, $\bm{W}^i$ is a $d \times 1$ weight matrix, and $\bm{b}^i$ is a one-dimensional bias parameter.
The hypernetwork $h$ customizes both the feature aggregation process and the final decision according to each patient's clinical data.
Specifically, given the tabular data $\bm{T}^i$, the hypernetwork outputs 
(1) the Tabular-conditioned Transformation Parameters (TCTP) $\bm{v}^i$, which modify the aggregation token to change the model’s focus on image instances, and (2) the patient-specific classifier weights and biases $(\bm{W}^i, \bm{b}^i)$, which adjust the decision criterion for the final prediction:
\begin{equation}
    \bm{v}^i, \bm{W}^i, \bm{b}^i = h(\bm{T}^i).
\end{equation}
Here, $\bm{v}^i$ is a $d$-dimensional vector with the same dimension as the aggregation token, $\bm{W}^i \in \mathbb{R}^{d\times1}$, and $\bm{b}^i \in \mathbb{R}$.
In this way, the model can adapt both its attention to image regions and its decision boundary based on patient-specific conditions.

\noindent
{\bf Adaptive aggregation.}
In this step, the aggregation process is adaptively adjusted for each patient by applying the Tabular-conditioned Transformation Parameters (TCTP) $\bm{v}^i$ estimated by the hypernetwork.

Given a bag $\mathcal{B}^i = \{\bm{x}_{j}^i\}_{j=1}^{|\mathcal{B}^i|}$, the instance feature extractor $f$ encodes each slice image as an instance feature $\bm{e}_j^i = f(\bm{x}_j^i)$. 
To condition the aggregation on patient information, the TCTP $\bm{v}^i$ is added to the learnable aggregation token $\bm{a}$, yielding a patient-specific aggregation token:
\begin{equation}
\tilde{\bm{a}}^{i} = \bm{a} + \bm{v}^i.
\end{equation}
The adapted token $\tilde{\bm{a}}^{i}$ and the instance features $\{\bm{e}_{j}^i\}_{j=1}^{|\mathcal{B}^i|}$ are then fed into the Adaptive Aggregation Transformer to obtain the bag-level feature:
\begin{equation}
\bm{q}^i = \text{AdAgFormer}(\tilde{\bm{a}}^{i}, \bm{e}_{1}^i, \ldots, \bm{e}_{|\mathcal{B}^i|}^i).
\end{equation}
Within the transformer, the adapted aggregation token guides the attention toward clinically important instances, enabling patient-specific feature aggregation.

Finally, the bag-level feature $\bm{q}^i$ is classified using the patient-specific classifier parameters $(\bm{W}^i, \bm{b}^i)$ generated by the hypernetwork:
\begin{equation}
\hat{Y}^i = \bm{q}^i \bm{W}^i + \bm{b}^i,
\end{equation}
and the binary cross-entropy loss is computed as
\begin{equation}
L^i = \text{BCE}(\hat{Y}^i, Y^i).
\end{equation}

\section{Experiments}
\noindent
{\bf Dataset.}
Since estimating the necessity of using debulking device is a novel task, no public datasets are available. 
% Therefore, we used a private dataset collected from an anonymous hospital, consisting of paired CT image and tabular data representing patient information.
Therefore, we used a private dataset collected from Kyushu University–affiliated hospitals, where each CT image is paired with tabular data representing patient information.
% The dataset consists of 113,594 images collected from 493 patients with patient-level necessity of using debulking devices labels. The number of images in the bag ranges from 9 to 635, with various bags, and the average bag size is 230. 
The dataset consists of CT images collected from 493 patients, each labeled with the patient-level necessity of using debulking device. The number of cross-sectional vessel images corresponding to each CT scan in a bag ranges from 9 to 635, with an average bag size of 230.
The tabular data comprise 20 items, including age, sex, the presence or absence of HF, LVEF, and other clinical factors.
Each method is evaluated with 5-fold cross-validation using a 3:1:1 split for training, validation, and test data.

\noindent
{\bf Implementation details and evaluation metrics.}
The method was implemented in PyTorch~\cite{Paszke2019PyTorchAI}. 
ResNet18~\cite{he2016deep} pre-trained on ImageNet~\cite{deng2009imagenet} was used as the instance feature extractor $f$, 
and the hypernetwork $h$ was a three-layer MLP with three linear heads for predicting $\bm{v}^i$, $\bm{W}^i$, and $\bm{b}^i$, 
initialized following Chang et al.~\cite{Chang2020Principled}. 
Training used the Adam optimizer~\cite{adam} with a learning rate of $3\times10^{-6}$, batch size 16, and early stopping (patience 50). 
% To implement our method, we used PyTorch.
% As the instance feature extractor $f$, we used ResNet18 pre-trained on the ImageNet dataset~. 
% The network was trained using the Adam optimizer, with a learning rate of $3e-6$, $\mathrm{epoch}=2000$, a mini-batch size$=16$, and early stopping$=50$ patients. 
% To address the class imbalance in the data, oversampling based on the number of patient-level labels was performed during training.
% To implement our method, we used PyTorch~\cite{Paszke2019PyTorchAI}.
% As the instance feature extractor $f$, we used ResNet18~\cite{he2016deep} pre-trained on the ImageNet dataset~\cite{deng2009imagenet}. 
% The network was trained using the Adam optimizer~\cite{adam}, with a learning rate of $3e-6$, $\mathrm{epoch}=2000$, a mini-batch size$=16$, and early stopping$=50$ patients. 
% To address the class imbalance in the data, oversampling based on the number of patient-level labels was performed during training.
% Each method is evaluated with 5-fold cross-validation.

% As evaluation metrics, we used the F1 score, which is robust to class imbalance, and the AUC, which provides a threshold-independent evaluation.

\begin{table}
    \centering
        \caption{{\bf Comparison with unimodal and multimodal MIL methods.} ``Modal'' indicates the number of data modalities used in each method.
        }
        \vspace{1mm}
        \scalebox{0.88}{
        \begin{tabular}{c|c||ccc}  \hline
           Method  &  Modal&F1 & AUC\\ \hline \hline
           TableMLP &Uni&0.464&0.578\\
        Output+Max~\cite{wang2018revisiting} &Uni&0.474&0.606\\
            Feature+Mean~\cite{Ilse2020DeepMI}&Uni&0.508&0.642\\
            Feature+Max~\cite{Ilse2020DeepMI} &Uni&0.479&0.549\\
            Feature+Attention~\cite{ilse2018attention}&Uni&0.498&0.659\\
            Feature+Transformer&Uni&0.544&0.667\\ \hdashline
            Concat &Multiple&0.558&0.688\\
            Gated Attention fusion~\cite{chen2020pathomic} &Multiple&0.547&0.686\\
            MultimodalTransformer~\cite{xu2023multimodal}&Multiple&0.521&0.676\\
            M3IFusion~\cite{li2021multi} &Multiple&0.517&0.667\\
            HEALNet~\cite{hemker2024healnet} &Multiple&0.492&0.642\\
                \rowcolor{gray!15}
                \rowcolor{gray!15}
            Ours &Multiple&{\bf 0.570}&{\bf 0.710}\\ \hline
            % Ours &Multiple&{\bf 0.573}&{\bf 0.696}\\ \hline
        \end{tabular}
        }
        \vspace{-4mm}
        \label{tab:comparison_w_unimodalMIL}
\end{table}

% \begin{table}
%     \centering
%         \caption{{\bf Comparison with unimodal MIL methods.} Best performances are bold.
%         }
%         \scalebox{0.9}{
%         \begin{tabular}{c||ccc}  \hline
%            Method  &  F1 & AUC\\ \hline \hline
%         Output+Max~\cite{wang2018revisiting} &0.474&0.606\\
%             Feature+Mean~\cite{Ilse2020DeepMI}&0.508&0.642\\
%             Feature+Max~\cite{Ilse2020DeepMI} &0.479&0.549\\
%             Feature+Attention~\cite{ilse2018attention}&0.498&0.659\\
%             Feature+Transformer&0.544&0.667\\
%                 \rowcolor{gray!15}
%             Ours &{\bf 0.573}&{\bf 0.696}\\ \hline
%         \end{tabular}
%         }
%         \vspace{-13mm}
%         \label{tab:comparison_w_unimodalMIL}
% \end{table}

% \begin{table}
%     \centering
%         \caption{{\bf Comparison with multimodal fusion methods, including those designed for MIL.} Best performances are bold.
%         }
%         \scalebox{0.9}{
%         \begin{tabular}{c||ccc}  \hline
%            Method  &  F1 & AUC\\ \hline \hline
%             Concat &0.558&0.688\\
%             Gated Attention fusion~\cite{chen2020pathomic} &0.547&0.686\\
%          MultimodalTransformer~\cite{xu2023multimodal}&0.521&0.676\\
%             M3IFusion~\cite{li2021multi} &0.517&0.667\\
%             HEALNet~\cite{hemker2024healnet} &0.492&0.642\\
%                 \rowcolor{gray!15}
%             Ours &{\bf 0.573}&{\bf 0.696}\\ \hline
%         \end{tabular}
%         }
%         \vspace{0mm}
%         \label{tab:comparison_w_multiimodal}
% \end{table}

\noindent
{\bf Performance evaluation.}
% To demonstrate the effectiveness of utilizing tabular data as an additional modality to image data in our proposed method and to highlight the difficulty of estimating the necessity of using debulking devices, we compare our method with five unimodal MIL methods that perform training and inference using only image data.
To demonstrate the effectiveness of utilizing multimodal data in the proposed method, we compare it with a tabular classification method and five unimodal MIL methods.
1) ``TableMLP'' classifies tabular data using an MLP.
2) ``Output+Max''~\cite{wang2018revisiting}, 3) ``Feature+Mean'', and 4) ``Max''~\cite{Ilse2020DeepMI} aggregate instance-level classification scores or features using max or mean pooling.
% 1) ``Output+Max''~\cite{wang2018revisiting} aggregates the instance-level classification scores using max pooling.
% 2) to 5) are feature aggregation methods that have become mainstream in recent MIL studies.
% ``Feature+Mean'' and ``Max'' aggregate instance features using mean pooling and max pooling, respectively~\cite{Ilse2020DeepMI}.
5) ``Feature+Attention''~\cite{ilse2018attention} and 6) ``Transformer'' aggregate instance features through weighted summation using a traditional attention mechanism or the self-attention mechanism.
% of the transformer.

The upper section of Table~\ref{tab:comparison_w_unimodalMIL} shows the bag-level estimation performance of the comparison methods.
The proposed method outperformed all conventional unimodal MIL approaches.
``TableMLP'' shows lower performance than image-based methods due to the limited information in the tabular data.
% ``Feature+Transformer'' achieves higher performance than other conventional methods by utilizing the self-attention mechanism; however, its performance remains limited because it cannot perform patient-specific adaptive aggregation.
``Feature+Transformer'' outperforms conventional methods by leveraging self-attention, but its static aggregation limits performance.
The superior results of the proposed method demonstrate the effectiveness of patient-specific adaptive aggregation based on tabular data.
% This result demonstrates the effectiveness of the proposed method in performing adaptive aggregation for each patient.

% Furthermore, the performance of all methods remains below a macro-F1 score of 0.6 and an AUC of 0.7, indicating that the task of estimating the necessity of using debulking devices is highly challenging.

% The performance of all methods remains below a macro-F1 score of 0.6 and an AUC of 0.7, indicating that estimating the necessity of using debulking devices is a highly challenging task.

\begin{table}
    \centering
        \caption{{\bf Ablation experiment.} ``Classifier'' and ``TCTP'' denote the modules generated by the hypernetwork.  
        }
                \scalebox{0.8}{
        \begin{tabular}{c|c|c||cc}  \hline
           Method&Classifier&TCTP& F1 & AUC\\ \hline \hline
           Feature+Transformer &&&0.544&0.667\\
            Ours w/o TCTP &$\checkmark$&&0.557&0.692\\
            \rowcolor{gray!15}
            Ours  &$\checkmark$&$\checkmark$&{\bf 0.570}&{\bf 0.710}\\ \rowcolor{gray!15}\hline
        \end{tabular}
                }
        \vspace{-2mm}
        \label{tab:ablation}
\end{table}

% \noindent
% {\bf Comparisons with Multimodal Fusion Methods.}
To evaluate the effectiveness of using tabular data as a condition for feature aggregation rather than for feature fusion, we compared our method with five multimodal fusion methods.
% 1) ``Concat'' concatenates image features obtained from a transformer-based MIL model with tabular features extracted by an MLP, and then inputs the fused features into a classifier to obtain the final prediction.
% 2) ``Gated Attention fusion''~\cite{chen2020pathomic} performs fusion by first weighting the features of each modality using a gated attention mechanism and then combining them through a Kronecker product.
7) ``Concat'' and 8) ``Gated Attention Fusion''~\cite{chen2020pathomic} fuse the aggregated bag-level features obtained from “Feature+Transformer” with a tabular feature vector using concatenation or a weighted Kronecker product,
9) to 11) are the multimodal MIL methods.
9) ``MultimodalTransformer''~\cite{xu2023multimodal} inputs instance and tabular feature vectors into a transformer, where they are aggregated using an aggregation token.
``M3IFusion''~\cite{li2021multi} concatenates instance features with tabular features and then performs aggregation using traditional attention-based MIL.
``HEALNet''~\cite{hemker2024healnet} is a state-of-the-art method that densely shares features between the image and tabular using cross-attention.

% Table~\ref{tab:comparison_w_multiimodal} shows the bag-level estimation performance of the comparison methods.
% % These results demonstrate that the proposed method outperforms conventional multimodal fusion methods.
% These results indicate that, while the conventional methods exhibit poor performance, the proposed method demonstrates superior performance.

The lower section of Table~\ref{tab:comparison_w_unimodalMIL} shows that while conventional methods perform poorly, the proposed method achieves superior results. In this task, incorporating tabular information during aggregation is crucial, as ``Concat'' and ``Gated Attention Fusion,'' which fuse tabular features after feature aggregation in conventional transformer-based methods, show lower performance. Moreover, conventional multimodal MIL approaches assume comparable information across modalities and perform dense feature sharing, leading to reduced performance in this task where tabular data contain far fewer variables and less information than images. In contrast, the proposed method enhances image feature aggregation by conditioning on tabular data and remains unaffected by the information imbalance between modalities.

\begin{figure}
      \centering
        \includegraphics[width=1.0\linewidth]{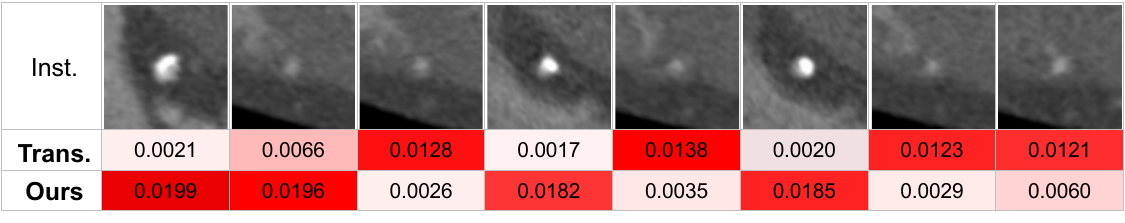}
        \vspace{-7mm}
        \caption{{\bf  Example of instances and their corresponding attention scores} estimated by ``Feature+Transformer'' and the proposed ``HyperAdAgFormer.'' 
        \label{fig:attn_vis}}
        \vspace{-3mm}
        
\end{figure}

% Table~\ref{tab:ablation} shows the results of the ablation study. ``Feature+Transformer'' represents a transformer-based unimodal MIL method, while ``Ours w/o TCTP'' removes the TCTP from the proposed method and uses the hypernetwork to generate only the classification layer from the tabular data.
% These results demonstrate the importance of not only generating the classifier based on tabular data but also adaptively modifying the feature aggregation.
Table~\ref{tab:ablation} summarizes the ablation results. ``Feature+\allowbreak Transformer'' denotes a transformer-based unimodal MIL method, while ``Ours w/o TCTP'' removes the TCTP from the proposed method and uses the hypernetwork only to generate the classifier from tabular data. These results highlight the importance of not only generating the classifier from tabular data but also adaptively modifying feature aggregation.

% To demonstrate the effectiveness of each module of HyperAdAgFormer, we compare our method with two ablation variants: ``TableMLP,'' which performs classification using only tabular data through an MLP, and ``Feature+Transformer,'' which aggregates instance features using a transformer without performing patient-specific adaptive aggregation.
% Table~\ref{tab:ablation} shows that using only tabular data results in a significant performance drop, likely due to its limited information content compared to image data.
% % The results in Table~\ref{tab:ablation} show that estimation using only tabular data exhibits a significant performance drop compared to methods that incorporate image data. This is likely because the tabular data contain relatively little information.
% % Both ``Feature+Transformer'' and the proposed HyperAdAgFormer show substantial performance improvements by utilizing image data. Furthermore, HyperAdAgFormer achieves additional performance gains by leveraging tabular data as a conditioning factor to enable patient-specific adaptive aggregation.
% ``Feature+Transformer'' achieves higher performance than ``TableMLP'' because it utilizes image data.
% The proposed HyperAdAgFormer further outperforms ``Feature+Transformer,'' demonstrating that adaptive aggregation based on tabular data is more effective than a static aggregation strategy.

To demonstrate the proposed method’s adaptive attention estimation capability, Fig.~\ref{fig:attn_vis} shows examples from ``Feature+Transformer'' and the proposed method for a device-treated patient with a high LVEF of 68.5\% (normal: 55–70\%) and no history of heart failure. As the calcification in this patient was relatively small, ``Feature+Transformer'' largely ignored these regions, whereas the proposed method, guided by tabular data reflecting good cardiac function, attended even to small calcified areas.
% To demonstrate the proposed method’s  estimate attention weights capability, Fig.~\ref{fig:attn_vis} shows examples of attention weights estimated by “Feature+Transformer” and the proposed “HyperAdAgFormer.” The results show that our method assigns greater attention to large calcified regions. Given the patient’s high LVEF (69\%), indicating good cardiac function, the method appears to aggregate features by focusing more precisely on calcified regions based on Tabular.
% To demonstrate the proposed method’s ability to estimate attention weights in adaptive aggregation, Fig.~\ref{fig:attn_vis} presents examples of attention weights estimated for instances by ``Feature+Transformer'' and the proposed ``HyperAdAgFormer.''
% This result indicates that the proposed method assigns greater attention to large calcified regions. Since the patient has a Left Ventricular Ejection Fraction (LVEF) of 69\%, indicating favorable cardiac function, it can be inferred that the proposed method aggregates features by focusing more precisely on the calcified regions based on the patient’s clinical information.

% To show the effectiveness 

% a

% \vspace{30mm}

\section{Conclusion}
% This study presents the first attempt to estimate the necessity of debulking devices from CT images by formulating the task as a MIL problem that predicts a patient-level label from multiple vascular slices. The challenge lies in that surgeons decide which calcifications to prioritize based on tabular data reflecting patient-specific factors. To address this, we propose HyperAdAgFormer, a hypernetwork-based adaptive aggregation transformer that adaptively adjusts the feature aggregation strategy for each patient using tabular data. Experimental results confirm the effectiveness of the proposed method.
This study presented the first attempt to estimate the necessity of debulking device from CT images by formulating it as a MIL problem predicting a patient-level label from multiple vascular slices.
The challenge of this task lies in the fact that surgeons determine which calcifications to prioritize by considering tabular data reflecting patient-specific factors.
To address this issue, we proposed a Hypernetwork-based Adaptive Aggregation Transformer (HyperAdAgFormer), which adaptively adjusts the feature aggregation strategy for each patient based on tabular data through a hypernetwork.
Experimental results demonstrated its effectiveness.
In future work, we intend to analyze and extend our approach to handle a large amount of missing values in tabular data.

% \noindent
\section{Compliance with ethical standard}
This study is approved by the Kyushu University Ethics Committee for Observational Clinical Research (22059, M22141). 

% \noindent
\section{Acknowledgments}
This work was supported by  JST BOOST, Japan Grant Number JPMJBS2406, SIP-JPJ012425, ASPIRE Grant Number JPMJAP2403.

% \vspace{50mm}

% \section{Acknowledgments}
% \label{sec:acknowledgments}

\bibliographystyle{IEEEbib}
\bibliography{strings,refs}

\end{document}